\documentclass[runningheads]{llncs}

\usepackage{eccv}

\usepackage{eccvabbrv}

\usepackage{graphicx}
\usepackage{booktabs}
\usepackage{enumitem}
\usepackage{multirow}
\usepackage{pifont}
\usepackage{bm}
\usepackage[table]{xcolor}
\usepackage{xcolor}
\definecolor{blueToken}{RGB}{31, 119, 180}  
\definecolor{redToken}{RGB}{214, 39, 40}   
\usepackage[accsupp]{axessibility}  
\usepackage{bbding} 

\usepackage{hyperref}

\usepackage{orcidlink}

\begin{document}

\title{Reasoning with Memory: A Temporal Granularity-Adaptive Framework for Training-Free Long Video Understanding} 

\titlerunning{Reasoning with Memory}


\author{Linghao Meng\inst{1,2}\textsuperscript{*} \and
Qiankun Li\inst{3}\textsuperscript{*} \and
Junyuan Mao\inst{1} \and
Pujin Liao\inst{4} \and
Zhicheng He\inst{1} \and
Enbo Zhang\inst{5} \and
Kun Wang\inst{3} \and
Yang Liu\inst{3} \and
Huazhu Fu\inst{2}\textsuperscript{(\Envelope)} \and
Yueming Jin\inst{1}\textsuperscript{(\Envelope)}}

\authorrunning{L. Meng, Q. Li et al.}

\institute{National University of Singapore, Singapore\\
\email{menglinghao25@u.nus.edu; ymjin@nus.edu.sg}
\and
Institute of High Performance Computing, Agency for Science, Technology and Research (A*STAR), Singapore\\
\email{hzfu@ieee.org}
\and
Nanyang Technological University, Singapore\\
\email{cs-qiankun.li@ntu.edu.sg}
\and
University of Science and Technology of China, Anhui, China \and
Jilin University, Jilin, China\\
}

\maketitle


\begin{abstract}
  While Multimodal Large Language Models (MLLMs) demonstrate superior generalization in fundamental video tasks, restricted context windows limit their long video understanding. To accommodate this constraint, models typically resort to keyframe selection. However, uniform sampling or static query-guided selection often overlooks critical temporal context, failing to adapt to the varying query temporal granularities. In this paper, we propose \textbf{ReMem}, a temporal granularity-adaptive keyframe selection framework for training-free LongVideoQA. ReMem introduces a \textbf{dual-level} memory-augmented adaptation. At the query level, \textbf{Memory-Driven Question Parsing} leverages LLM long-term memory to decode question temporal granularity and extract semantic entities. At the video level, \textbf{Synergistic Dual-Semantic Frame Alignment} exploits intrinsic structural memory to align frames with query semantics, guiding \textbf{Structure-Aware Dynamic Frame Routing} to cluster events and optimally distribute sampling budgets. By explicitly preserving temporal information with memory mechanisms, ReMem suppresses redundancy and empowers MLLMs to perform robust multi-granular video reasoning. Evaluations across four popular LongVideoQA benchmarks using three MLLMs demonstrate highly efficient, state-of-the-art zero-shot performance; notably, LLaVA-Video with ReMem reaches 54.5\% (+12.3\%) on LVBench and 67.1\% (+8.2\%) on LongVideoBench. Code is available at \href{https://github.com/jinlab-imvr/ReMem}{ReMem}.
  \keywords{Training-free LongVideoQA \and Memory Mechanism \and Temporal Granularity}
\end{abstract}

\begingroup
\renewcommand{\thefootnote}{}
\footnotetext{
	\textsuperscript{*} Equal contribution.\\
	\textsuperscript{\Envelope} Corresponding authors.
    \vspace{-0.6em}
}
\endgroup

\section{Introduction}
\label{sec:intro}


Recently, Multimodal Large Language Models (MLLMs) have achieved remarkable breakthroughs in cross-modal reasoning by aligning visual encoders with LLMs~\cite{liu2023visual,zhu2023minigpt,bai2023qwen}. By integrating temporal modeling, this architecture extends to the video domain, adapting static perception for dynamic understanding~\cite{lin2024video,maaz2024video,li2025videochat,li2024llama,zhang2026bimm}. Consequently, MLLMs inherit robust instruction-following capabilities and demonstrate superior generalization across fundamental video tasks~\cite{jin2024chat,zhang2025collaboratively}.

Despite these advancements, standard MLLMs struggle with long video understanding tasks~\cite{fu2025video,zhou2024mlvu,wang2025lvbench}. The extensive frame counts exceed restricted context windows, precluding the processing of all frames~\cite{zhang2024long,song2024moviechat,zhang2025q}. Consequently, models often resort to sparse uniform sampling, which inadvertently omits crucial evidence and causes erroneous predictions~\cite{wang2023internvid,tang2025adaptive}. To address this, recent approaches focus on selecting the most discriminative keyframes within a limited frame budget~\cite{zhang2024simple,tang2025adaptive,li2025frameoracle,yu2023self}.

\begin{figure}[tb]
  \centering
  \includegraphics[width=\linewidth]{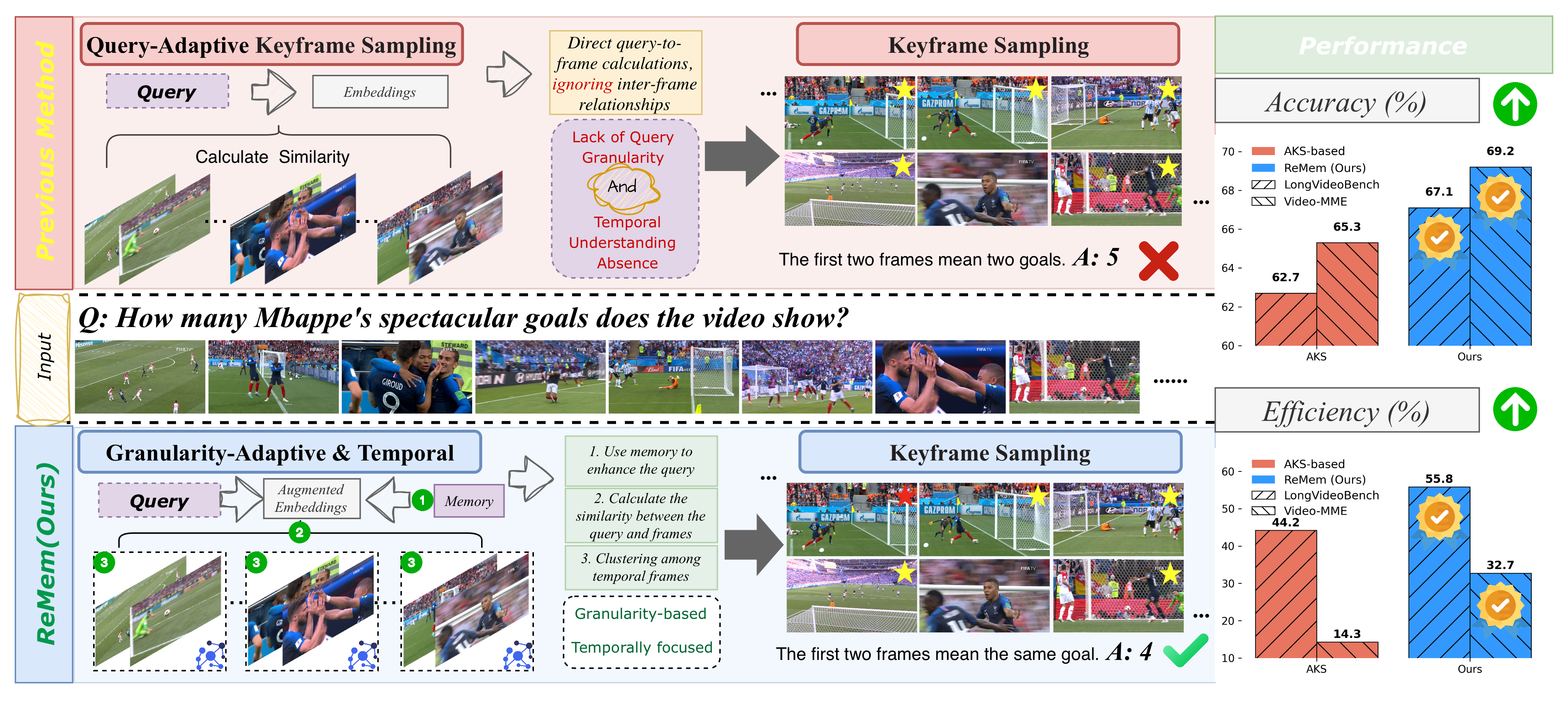}
  \caption{Query-adaptive methods evaluate frames in isolation and suffer from the absence of temporal understanding. Our \textbf{ReMem} overcomes this bottleneck with memory-augmented temporal granularity-adaptive sampling, achieving superior accuracy and efficiency for long video reasoning.}
  \label{fig:intro}
\vspace{-1em}
\end{figure}

However, current methods typically rely on static query-frame similarity for keyframe selection~\cite{ren2024timechat,tang2025adaptive,zhang2025q}, failing to adapt to the varying temporal granularities of different questions. As demonstrated in Fig.~\ref{fig:intro}, purely query-adaptive sampling tends to over-sample redundant frames from localized high-scoring segments. Consequently, while adequate for fine-grained queries, these approaches overlook essential temporal dependencies, resulting in suboptimal performance on tasks requiring continuous temporal reasoning.
Moreover, existing methods overlook the latent concepts embedded in questions, failing to enhance the textual signal-to-noise ratio for effective query-frame alignment.

To address these limitations in a training-free setting, the concept of \textbf{memory} serves as a dual-faceted mechanism to process and align both the query and the video. For the query, the inherent \textbf{long-term memory} of LLMs encapsulates rich cognitive priors, providing the intuition to analyze temporal granularity and make semantic associations~\cite{achiam2023gpt}. For the visual content, mining the video's inherent \textbf{structural memory} captures its continuous temporal evolution, effectively bridging the contextual gap between scattered frames.
Building upon this insight, we propose \textbf{ReMem} (\textit{\textbf{Re}asoning with \textbf{Mem}ory}), a temporal granularity-adaptive training-free keyframe selection framework for zero-shot LongVideoQA augmented by a memory mechanism.

ReMem formulates a memory-augmented dual-level adaptation framework, aligning question semantics with video visual and temporal dynamics.
At the \textit{query level}, a Reasoning LLM utilizes long-term memory to conduct \textbf{Memory-Driven Question Parsing}, evaluating the query temporal granularity and extracting key entities to guide visual retrieval. 
At the \textit{video level}, \textbf{Synergistic Dual-Semantic Frame Alignment} aligns frames with query semantics, guiding \textbf{Structure-Aware Dynamic Frame Routing} to cluster events and distribute budgets.
This memory-augmented strategy suppresses temporal redundancy and preserves causal structures, empowering MLLMs to perform robust multi-granular video reasoning.

We evaluate ReMem using three MLLMs across four LongVideoQA benchmarks, where it achieves state-of-the-art zero-shot performance with high efficiency.
On LVBench~\cite{wang2025lvbench} and LongVideoBench~\cite{wu2024longvideobench}, LLaVA-Video with ReMem reaches accuracies of 54.5\%(+12.3\%) and 67.1\%(+8.2\%), respectively.
These substantial gains are consistently observed across all evaluated MLLMs, highlighting ReMem as a generalizable framework that requires no parameter tuning.

The main contributions of this work are summarized as follows:
\begin{itemize}
    \item[\ding{182}] We propose \textbf{ReMem}, a training-free framework that performs temporal granularity-adaptive keyframe selection for long video understanding, enabling MLLMs to reason over long videos under strict context constraints.
    \item[\ding{183}] We introduce a \textbf{memory-driven dual-level adaptation mechanism} that leverages LLM long-term memory to parse query temporal granularity and employs structural video memory to align frames through structure-aware dynamic routing.
    \item[\ding{184}] Extensive experiments across three MLLM architectures and four popular LongVideoQA benchmarks demonstrate that ReMem achieves new state-of-the-art zero-shot performance.
\end{itemize}

\section{Related Work}

\subsection{Video Large Language Models}
Recent Video Large Language Models have witnessed significant advancements in visual-text alignment and spatial-temporal modeling~\cite{zhang2024llava,bai2025qwen3,li2024llava,liu2024llavanext,bai2025qwen2,wang2024internvideo2,zhang2025collaboratively,zhang2026bimm}. For instance, LLaVA-OneVision~\cite{li2024llava} and LLaVA-NeXT-QW2~\cite{liu2024llavanext} introduce AnyRes strategies and dynamic visual pooling to effectively process high-resolution frames across diverse scenarios. VideoLLaMA3~\cite{zhang2025videollama} further strengthens spatial-temporal representation through an advanced visual-language architecture, while Chat-UniVi-V1.5~\cite{jin2024chat} employs dynamic token merging to optimize computational efficiency without compromising performance.

To address the computational challenges of Long-Video Understanding, specialized architectures have emerged to decouple sequence length from memory cost. LongVA~\cite{zhang2024long} and LongVILA~\cite{chen2024longvila} significantly extend the context window by incorporating efficient spatial-temporal pooling and system-level training optimizations. Video-XL~\cite{shu2025video} compresses hour-scale videos into manageable KV caches utilizing latent visual summarization tokens. Similarly, LongVU~\cite{shen2024longvu} and Video-CCAM~\cite{fei2024video} introduce spatiotemporal context compression and causal cross-attention masking, effectively mitigating the heavy memory burden. Furthermore, Video-R1~\cite{feng2025video} and VideoChat-R1~\cite{li2025videochat-r1} integrate deductive reasoning traces into video comprehension, generating intermediate logical steps and enabling MLLMs to ground with complex temporal evidence and structured visual cues.

\subsection{Keyframe Selection}
The computational bottleneck of processing dense video tokens motivates the development of Keyframe Selection~\cite{zhao2023search,zhu2025focus,guo2025logic,tang2025adaptive,song2026ktv}, which aims to identify the most informative frames to represent the video content. These methods can be broadly categorized into training-based and training-free approaches.

Training-based methods typically learn dedicated modules to extract relevant evidence (e.g., Frame-Voyager~\cite{yu2024frame}, KeyVideoLLM~\cite{liang2024keyvideollm}, FrameOracle~\cite{li2025frameoracle}), or fine-tune lightweight LLMs as plug-and-play filters, such as Selector~\cite{hu2025m}. Alternatively, recent works like CoF~\cite{ghazanfari2025chain} and GenS~\cite{yao2025generative} integrate selection directly into the reasoning process by training Video-LLMs to generate temporal traces.

Training-free methods offer plug-and-play compatibility with off-the-shelf models, transitioning from static uniform sampling to query-adaptive strategies. For instance, AKS~\cite{tang2025adaptive} optimizes the balance between relevance and coverage, while Q-Frame~\cite{zhang2025q} and FlexSelect~\cite{zhang2025flexselect} dynamically adjust visual budgets through multi-resolution and context-aware probabilistic sampling respectively. Furthermore, BOLT~\cite{liu2025bolt} employs dynamic routing to match evidence with intent, and MDP3~\cite{sun2025mdp3} utilizes determinantal point processes for list-wise diversity. However, these approaches predominantly evaluate frames based on isolated static semantics, ignoring the continuous temporal dynamics and long-range causal dependencies required for complex long video reasoning.

\section{Method}

\begin{figure}[tb]
  \centering
  \includegraphics[width=\linewidth]{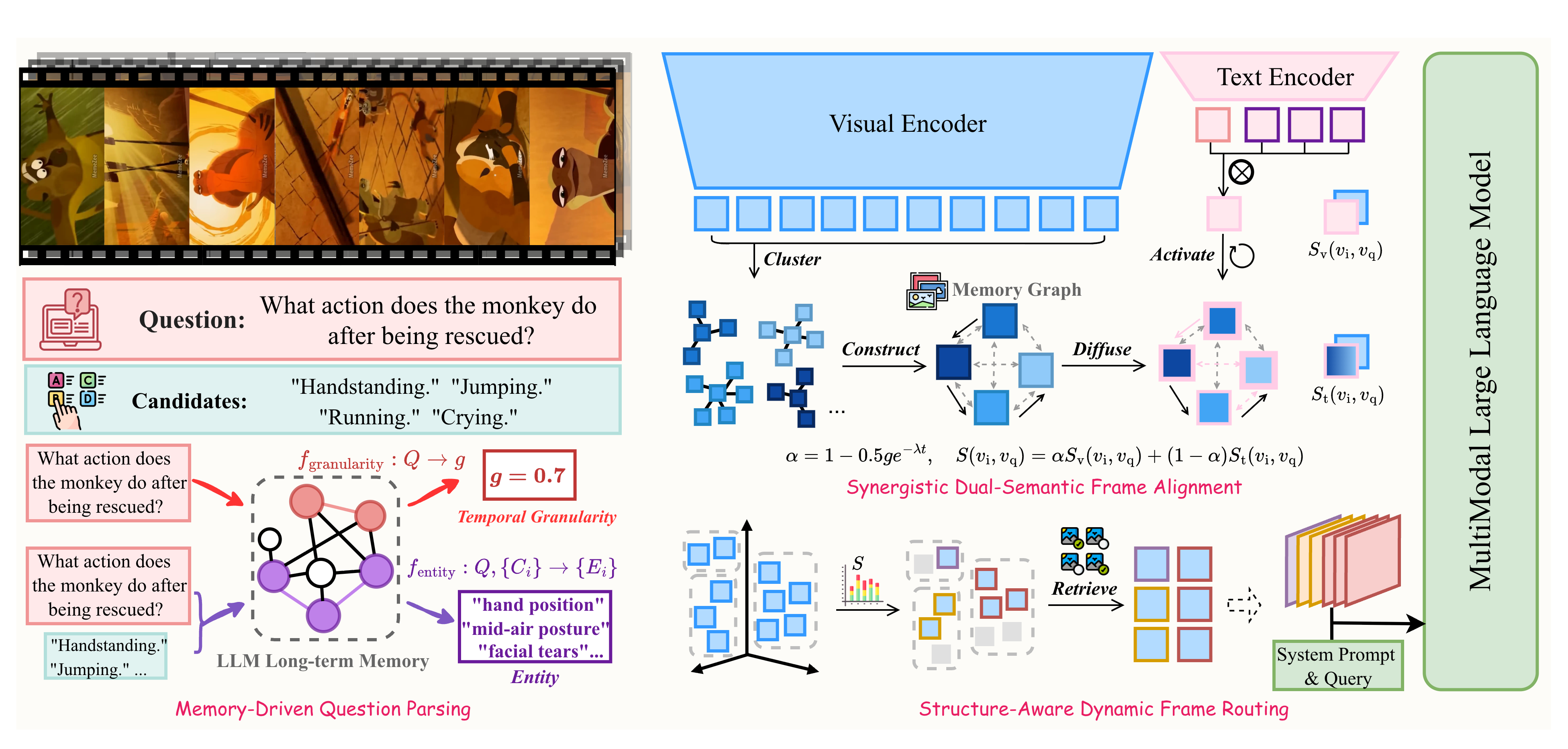}
  \caption{
        \textbf{Overview of ReMem.}
        ReMem performs dual-level adaptation for training-free LongVideoQA:
        \ding{182} \textbf{query-level adaptation}, where an LLM evaluates the question's temporal granularity and extracts entities; 
        \ding{183} \textbf{video-level adaptation}, where memory-augmented dual-semantic evidence assists in selecting keyframes.
        The pipeline consists of three modules: Memory-Driven Question Parsing,
        Synergistic Dual-Semantic Frame Alignment, and Structure-Aware Dynamic Frame Routing.
  }
  \label{fig:method}
\vspace{-1em}
\end{figure}

We propose ReMem, a training-free framework that adaptively selects the most informative frames for zero-shot Long Video Question Answering (LongVideoQA). 
As shown in Fig.~\ref{fig:method}, ReMem performs a comprehensive dual-level memory adaptation for long video understanding:
\ding{182} \textbf{query-level adaptation}, where a reasoning LLM utilizes long-term memory to evaluate the question's granularity and extract key entities; 
\ding{183} \textbf{video-level adaptation}, where keyframes are selected by memory-augmented visual evidence distillation.
The overall pipeline consists of three core modules: \textbf{Memory-Driven Question Parsing} ($\triangleright$ Section \ref{sec:g1}), \textbf{Synergistic Dual-Semantic Frame Alignment} ($\triangleright$ Section \ref{sec:g2}), and \textbf{Structure-Aware Dynamic Frame Routing} ($\triangleright$ Section \ref{sec:g3}).

\subsection{Memory-Driven Question Parsing}
\label{sec:g1}
\paragraph{Granularity Analysis.}
Previous training-free LongVideoQA approaches often select informative frames without considering the varying complex temporal granularity in question semantics. 
To address this, we formulate granularity analysis as a continuous reasoning process $f_{\text{granularity}}: Q \to g$, where an LLM draws upon its \textbf{long-term memory} to score the query $Q$, yielding $g \in [0, 1]$ that quantifies its temporal granularity.
Specifically, a lower $g$ denotes localized events requiring dense visual evidence from a narrow window, while a higher $g$ reflects overarching queries demanding long-range contextual integration (\eg, ``How many times is the sun visible in the video?'' $\to g=0.90$).

\paragraph{Entity Extraction.}
Beyond temporal granularity, raw queries often lack explicit visual cues for precise alignment. To bridge this gap, we formulate entity extraction as $f_{\text{entity}}: Q, \{C_i\} \to \{E_i\}$
, where the LLM leverages its \textbf{long-term memory} to extract concrete visual signatures from the question $Q$ and candidate options $\{C_i\}$ (\eg, ``How many Mbappe's spectacular goals does the video show?'', $\{\text{``6''}, \text{``5''}, \text{``4''}, \text{``3''}\} \to \{$``Mbappe scoring celebration'', ``soccer net shaking'', ``Mbappe's jersey number''$\}$).

\subsection{Synergistic Dual-Semantic Frame Alignment}
\label{sec:g2}
\paragraph{Query Enhancement.}
We extract the CLIP text embeddings $\bm{v}_q\!\in\!\mathbb{R}^{D}$ and $\bm{v}_e\!\in\!\mathbb{R}^{N \times D}$ for the question $Q$ and extracted entities $\{E_i\}$, respectively.
We dynamically inject $\bm{v}_e$ into $\bm{v}_q$ via a cross-attention residual stream:
\begin{equation}
\bm{v}_{anc} = \text{LN} \left( \bm{v}_q + \bm{W}_f \left[ \bm{v}_q \parallel \text{Softmax} \left( \frac{(\bm{W}_q \bm{v}_q)(\bm{W}_k \bm{v}_e)^\top}{\sqrt{D}} \right) (\bm{W}_v \bm{v}_e) \right] \right),
\label{eq:query_fusion}
\end{equation}
where $\text{LN}(\cdot)$ denotes Layer Normalization,
$\bm{W}_q, \bm{W}_k$, and $\bm{W}_v \in \mathbb{R}^{D \times D}$ are projection matrices,
$[\cdot \parallel \cdot]$ denotes the concatenation operation along the feature dimension,
and $\bm{W}_f \in \mathbb{R}^{2D \times D}$ is the fusion weight.
This yields $\bm{v}_{anc}$, a visually-grounded enhanced query anchor that boasts a significantly improved signal-to-noise ratio for subsequent frame retrieval.

\paragraph{Static Visual-Semantic Alignment.}
Let $\mathcal{I} = \{I_i\}_{i=1}^M$ denote the sequence of frames uniformly sampled from the raw video. For each frame $I_i$, we extract the CLIP visual embedding $\bm{v}_i \in \mathbb{R}^D$. 
To measure the semantic affinity while controlling scale and higher-order correlation, ReMem defines the static visual-semantic similarity as:
\begin{equation}
\mathcal{S}_{\text{v}}(\bm{v}_i, \bm{v}_{anc}) = \alpha \frac{(\bm{W}_s \bm{v}_{anc})^\top (\bm{W}_s \bm{v}_i)}{\|\bm{W}_s \bm{v}_{anc}\|_2 \, \|\bm{W}_s \bm{v}_i\|_2} + \beta \left( \frac{\bm{v}_{anc}^\top \bm{W}_c \bm{v}_i}{\sqrt{D}} \right)^2,
\label{eq:visual_semantic}
\end{equation}
where $\bm{W}_s, \bm{W}_c \in \mathbb{R}^{D \times D}$ denote the projection weight matrices inherited from the pre-trained CLIP model, $\bm{W}_s$ serves as a scale-controlling projection to alleviate the modality gap, and $\alpha, \beta$ are weighting coefficients governing the contribution of each term.

\paragraph{Memory Augmented Temporal-Semantic Alignment.}
To effectively capture cross-scene correlations and long-range dependencies, we construct a temporal-semantic memory graph $\mathcal{G} = (\mathcal{V}, \mathcal{E})$. 
We define the node set $\mathcal{V}$ to represent core visual anchor states, dynamically setting its size $K = \sqrt{M}$ for sub-linear scaling. 
Specifically, we partition the visual features $\{\bm{v}_i\}_{i=1}^M$ into $K$ clusters $\{C_k\}_{k=1}^K$ with centroids $\{\bm{\mu}_k\}_{k=1}^K$ via K-Means. 
To avoid abstract mean representations, each anchor is instantiated by the actual frame closest to its centroid, i.e., $\bm{V}_k = \arg\min_{\bm{v}_i \in C_k} \|\bm{v}_i - \bm{\mu}_k\|_2$. 
These instantiated nodes are chronologically ordered and stacked to form the memory feature matrix $\bm{V} = [\bm{V}_1, \dots, \bm{V}_K]^\top \in \mathbb{R}^{K \times D}$. 
Subsequently, we establish the edge set $\mathcal{E}$ by computing semantic weights $e_{sem}$ and assigning temporal weights $e_{tem}$ for local connectivity. 
The edge construction is mathematically formulated as:
\begin{equation}
    e_{sem}^{i,j} = \max(\bm{V}_i^\top \bm{V}_j, 0), \quad
    e_{tem}^{i,j} = \begin{cases} 1, & |i - j| \le 1 \\ 0, & \text{otherwise} \end{cases}.
\label{eq:memory_graph}
\end{equation}
To ensure stable energy propagation during graph reasoning, we construct the transition matrices $\bm{W}_{sem}, \bm{W}_{tem} \in \mathbb{R}^{K \times K}$ by explicitly applying $L_1$ normalization to the raw edge weights $e_{sem}$ and $e_{tem}$, respectively. 
For cross-scene reasoning, the enhanced semantic anchor $\bm{v}_{anc}$ initializes a seed probability distribution $\bm{p}^{0} \in \mathbb{R}^K$ across the anchor nodes, formulated as $\bm{p}^{(0)} = \text{Softmax}(\bm{V} \bm{v}_{anc})$. 
This activation energy diffuses through the graph via a random walk mechanism:
\begin{equation}
\bm{p}^{t+1} = (1 - \gamma) \Big( \alpha \bm{W}_{sem} + (1 - \alpha) \bm{W}_{tem} \Big) \bm{p}^{t} + \gamma \bm{p}^{0},
\label{eq:graph_reasoning}
\end{equation}
where $\gamma \in (0, 1)$ is the restart probability governing the diffusion depth, and $\alpha$ is a coefficient balancing content-driven attention and temporal inductive bias.
After 2 steps of diffusion, we obtain $\bm{p}^{2}$, which captures neighborhood interactions while preserving discriminative local topologies, thereby encoding the global structural importance of each anchor node. The temporal-semantic score is derived by:
\begin{equation}
\mathcal{S}_{\text{t}}(\bm{v}_i, \bm{v}_{anc}) = \mathbb{E}_{k \sim \bm{p}^{2}} \big[ \bm{v}_i^\top \bm{V}_k \big] = \sum_{k=1}^K \left( (\bm{v}_i^\top \bm{V}_k) p_k^{2} \right),
\label{eq:temporal_score}
\end{equation}
where $\mathbb{E}_{k \sim \bm{p}^{2}}$ denotes the expectation over the memory nodes guided by the propagated probabilities, and $p_k^{2}$ denotes the $k$-th scalar element of $\bm{p}^{2}$.

\paragraph{Granularity Guided Dual-Semantic Fusion.}
To comprehensively evaluate frame relevance, we formulate a joint similarity metric $\mathcal{S}$ that dynamically arbitrates between instantaneous spatial cues and global topological memory:
\begin{equation}
    \alpha = 1-0.5g e^{-\lambda t}, \quad \mathcal{S} = \alpha \mathcal{S}_{\text{v}} + (1 - \alpha) \mathcal{S}_{\text{t}},
\label{eq:final_score}
\end{equation}
where $\lambda$ is a decay constant and $t$ denotes the video duration.
Crucially, the balancing factor $\alpha$ is modulated by the query's temporal granularity $g$. A smaller $g$ indicating localized events yields a higher $\alpha$, prioritizing $\mathcal{S}_{\text{v}}$ to capture precise spatial evidence without temporal noise. Conversely, for a larger $g$ or longer video, $\alpha$ smoothly decays, elevating $\mathcal{S}_{\text{t}}$ to track evolving actions and long-range structural dependencies.
Ultimately, based on aggregated $\mathcal{S}$, we select the top $\textbf{N}$ frames to construct a synergistic dual-semantic \textbf{candidate pool}.

\subsection{Structure-Aware Dynamic Frame Routing}
\label{sec:g3}

\paragraph{Temporally-Coherent Event Clustering.}
To partition the N selected frames into $L$ coherent events, we jointly optimize the probabilistic assignment matrix $\bm{U} \in [0,1]^{N \times L}$, semantic centroids $\bm{c}_l$, and temporal centers $\tau_l$ by minimizing:
\begin{equation}
\min_{\bm{U}, \bm{c}, \bm{\tau}} \mathcal{L}_{\text{cluster}} = \sum_{i=1}^N \sum_{l=1}^L U_{i,l} \Big( \underbrace{1 - \bm{v}_i^\top \bm{c}_l}_{\textcolor{blueToken}{\text{semantic cost}}} + \beta \underbrace{(t_i - \tau_l)^2}_{\textcolor{redToken}{\text{temporal cost}}} \Big) - \epsilon \mathcal{H}(\bm{U}),
\label{eq:clustering}
\end{equation}
where $\mathcal{H}(\bm{U}) = -\sum_{i,l} U_{i,l} \log U_{i,l}$ encourages smooth transitions, $\epsilon$ controls cluster assignment softness, and $\beta$ regulates temporal compactness. Frames in the candidate pool are assigned to event clusters $\{G_1, \dots, G_L\}$ via $\arg\max_l U_{i,l}$.

\paragraph{Dual-Semantic Similarity-Guided Distribution.} 
To allocate the total frame budget $B$, we design a dynamic routing mechanism guided by the dual-semantic score $\mathcal{S}$. We define the information capacity $I_l$ and frame quota $m_l$ for each event $G_l$:
\begin{equation}
I_l = \sum_{i \in G_l} \mathcal{S}_i + \eta \log(1 + |G_l|), \quad m_l = B \frac{\exp( I_l / \theta )}{\sum_{j=1}^L \exp( I_j / \theta )},
\label{eq:allocation}
\end{equation}
where $\theta$ controls the distribution sharpness, and $\eta$ scales the penalty against redundant temporal sampling. $m_l$ is rounded to the nearest integer.

\paragraph{Temporal Synchronization and MLLM Integration.}
Based on $\mathcal{S}$, we retrieve the top $m_l$ frames from each fragment $G_l$. The gathered $B$ keyframes are monotonically sorted into a temporal sequence $\mathcal{V}_{\text{select}}$ and interleaved with the system prompt $\text{P}_{\text{sys}}$ and query $\text{Q}$. Finally, the joint representation $\mathcal{X} = \{ \text{P}_{\text{sys}}, \text{Q}, \mathcal{V}_{\text{select}} \}$ is fed into the MLLM for cross-modal reasoning.

\section{Experiments}

\begin{table*}[t]
\centering
\setlength{\abovecaptionskip}{6pt}
\setlength{\belowcaptionskip}{0pt}
\caption{Performance comparison on four long video understanding benchmarks. Our approach is evaluated across three MLLMs under fixed input frame settings. The best results are highlighted in \textbf{bold}. Methods denoted with $^*$ are training-based.}
\label{tab:main_results}
\resizebox{\textwidth}{!}{
\begin{tabular}{lccccccccc}
\toprule
\multirow{2}{*}{Model} & \multirow{2}{*}{LLM Size} & \multirow{2}{*}{\#Frames} & \multirow{2}{*}{LVBench} & \multirow{2}{*}{MLVU} & \multirow{2}{*}{LongVideoBench} & \multicolumn{4}{c}{Video-MME} \\
\cmidrule(lr){7-10}
& & & & & & Overall & Short & Medium & Long \\
& Avg. Video Duration & & 68min & 12min & 12min & 17min & 1.3min & 9min & 41min \\
\midrule
\rowcolor{gray!25} \multicolumn{10}{l}{\textbf{Video-LLMs:}} \\
Video-LLaVA~\cite{lin2024video} & 7B & 8 & - & 47.3 & 39.1 & 39.9 & 45.3 & 38.0 & 36.2 \\
Qwen-VL~\cite{bai2023qwen} & 7B & 8 & - & - & - & 41.1 & 46.9 & 38.7 & 37.8 \\
VideoChat2~\cite{li2024mvbench} & 7B & 8 & - & 44.5 & 39.3 & 39.5 & 48.3 & 37.0 & 33.2 \\
Chat-UniVi-V1.5~\cite{jin2024chat} & 7B & 8 & - & - & - & 40.6 & 45.7 & 40.3 & 35.8 \\
VideoLLaMA2~\cite{cheng2024videollama} & 7B & 8 & - & - & - & 47.9 & 56.0 & 45.4 & 42.1 \\
LLaVA-NeXT-QW2~\cite{liu2024llavanext} & 7B & 8 & - & - & - & 49.5 & 58.0 & 47.0 & 43.4 \\
LongVILA~\cite{chen2024longvila} & 8B & 128 & - & - & - & 49.2 & 60.2 & 48.2 & 38.8 \\
LongVA~\cite{zhang2024long} & 7B & 128 & - & - & - & 52.6 & 61.1 & 50.4 & 46.2 \\
Video-XL~\cite{shu2025video} & 7B & 128/256 & - & 64.9 & - & 55.5 & 64.0 & 53.2 & 49.2 \\
LLaVA-OneVision~\cite{li2024llava} & 7B & - & - & 64.7 & 56.3 & 58.2 & - & - & - \\
Video-CCAM~\cite{fei2024video} & 9B & 96 & - & 58.5 & - & 50.3 & 61.9 & 49.2 & 39.6 \\
LongVU~\cite{shen2024longvu} & 7B & 1fps & - & 65.4 & - & 60.9 & 64.7 & 58.2 & 59.5 \\
VideoChat-R1~\cite{li2025videochat-r1} & 7B & 32 & 34.3 & - & 49.1 & - & - & - & 46.2 \\
Video-R1~\cite{feng2025video} & 7B & 32 & 35.3 & 45.4 & 52.7 & - & - & - & 48.2 \\
VideoLLaMA3~\cite{zhang2025videollama} & 7B & 16 & - & 50.9 & 56.1 & 61.2 & - & - & - \\
\midrule
\rowcolor{gray!25} \multicolumn{10}{l}{\textbf{Fixed input frames:}} \\
LLaVA-Video~\cite{zhang2024llava} & 7B & 64 & 42.2 & 70.8 & 58.9 & 64.4 & 76.6 & 62.5 & 54.2 \\
\enspace + AKS~\cite{tang2025adaptive} & 7B & 64 & - & - & 62.7 & 65.3 & 76.9 & 65.0 & 54.1 \\
\enspace + GenS~\cite{yao2025generative}$^*$ & 7B & 54 & - & 73.4 & 63.3 & - & - & - & - \\
\enspace + BOLT~\cite{liu2025bolt} & 7B & 8 & - & - & - & 58.6 & 70.4 & 55.7 & 49.9 \\
\enspace + FlexSelect-Lite~\cite{zhang2025flexselect} & 7B & 64 & 52.2 & 71.8 & 60.7 & 68.3 & - & - & 58.3 \\
\textbf{\enspace + ReMem (ours)} & \textbf{7B} & \textbf{64} & \textbf{54.5} & \textbf{77.3} & \textbf{67.1} & \textbf{69.2} & \textbf{77.6} & \textbf{68.2} & \textbf{61.8} \\
\midrule
Qwen2-VL~\cite{wang2024qwen2} & 7B & 32 & 41.5 & 56.9 & 55.5 & 57.6 & 69.9 & 56.4 & 46.5 \\
\enspace + AKS~\cite{tang2025adaptive} & 7B & 32 & - & - & 60.5 & 59.9 & - & - & - \\
\enspace + GenS~\cite{yao2025generative}$^*$ & 7B & 54 & - & 66.9 & 58.7 & - & - & - & - \\
\enspace + Q-Frame~\cite{zhang2025q} & 7B & 32 & - & 65.4 & 58.4 & 58.3 & 69.4 & 57.1 & 48.3 \\
\enspace + Selector~\cite{hu2025m}$^*$ & 7B & 32 & - & - & - & 58.7 & 69.6 & 54.1 & 51.9 \\
\textbf{\enspace + ReMem (ours)} & \textbf{7B} & \textbf{32} & \textbf{51.7} & \textbf{72.8} & \textbf{64.1} & \textbf{65.0} & \textbf{73.6} & \textbf{65.7} & \textbf{55.8} \\
\midrule
Qwen3-VL~\cite{bai2025qwen3} & 8B & 32 & 42.7 & 63.6 & 58.6 & 64.1 & 73.4 & 63.4 & 55.4 \\
\enspace + FrameOracle~\cite{li2025frameoracle}$^*$ & 8B & 20.9 & - & 62.9 & 64.0 & 67.3 & - & - & - \\
\textbf{\enspace + ReMem (ours)} & \textbf{8B} & \textbf{32} & \textbf{53.3} & \textbf{77.6} & \textbf{64.4} & \textbf{68.8} & \textbf{77.7} & \textbf{68.2} & \textbf{60.5} \\
\bottomrule
\end{tabular}
}
\end{table*}

\subsection{Experimental Setup}

\noindent\textbf{\ding{68} Evaluation Benchmarks and Metric.} We evaluate our method on four public long video benchmarks: Video-MME~\cite{fu2025video}, LongVideoBench~\cite{wu2024longvideobench}, MLVU~\cite{zhou2024mlvu}, and LVBench~\cite{wang2025lvbench}, which assess distinct dimensions of prolonged multimodal reasoning. To focus on visual understanding and weaken MLLM's language bias, tasks are formulated as subtitle-free multiple-choice question answering. Following standard protocols, we report the answer selection accuracy as the metric.

\noindent\textbf{\ding{68} Implementation Details.} We instantiate our framework on three foundational Video-LLMs: LLaVA-Video-7B-Qwen2~\cite{zhang2024llava} (frame budget $B$ is set to 64), Qwen2-VL-7B-Instruct~\cite{wang2024qwen2} and Qwen3-VL-8B-Instruct~\cite{bai2025qwen3} ($B$ is set to 32). We employ GPT-4o~\cite{achiam2023gpt} as the reasoning LLM and OpenAI’s CLIP-L-14~\cite{radford2021learning} as the encoder to extract visual and textual representations. To alleviate computational overhead while maintaining video information completeness, we construct the initial frames sequence $\mathcal{I}$ by uniformly sampling at 1 fps. During subsequent selection phases, $N$ is empirically set to 150, and $\lambda$ is fixed at $10^{-4}$. All experiments are zero-shot, training-free, and performed on 8$\times$NVIDIA H200 GPUs.

\subsection{Main Results}
As summarized in Table \ref{tab:main_results}, we comprehensively evaluate our proposed framework against state-of-the-art Video-LLMs and existing frame selection strategies. The quantitative results unequivocally demonstrate the superiority of ReMem, which we analyze across three critical dimensions.

\noindent\textbf{Consistent High Performance.} 
Our framework establishes a new SOTA across multiple challenging benchmarks, demonstrating robust generalizability independent of LLM size or architecture. When integrated into the $7$B LLaVA-Video backbone, ReMem yields a substantial $6.5\%\uparrow$ gain on MLVU~\cite{zhou2024mlvu}. Performance leaps are also striking with the $8$B Qwen3-VL backbone, where we observe a compelling $10.6\%\uparrow$ absolute improvement on LVBench~\cite{wang2025lvbench}, alongside a significant jump on MLVU~\cite{zhou2024mlvu} (\(63.6\%\!\rightarrow77.6\%\)). These consistent improvements validate our temporal granularity-adaptive framework's effectiveness.

\noindent\textbf{Efficacy of the Training-Free Paradigm.} 
A hallmark advantage of ReMem is its ability to achieve exceptional performance without the need for expensive, domain-specific fine-tuning. Strikingly, our zero-shot formulation consistently surpasses recent strongly supervised frame-selection baselines. To illustrate, while the training-intensive method GenS~\cite{yao2025generative} achieves a respectable $66.9\%$ on MLVU~\cite{zhou2024mlvu} using Qwen2-VL, ReMem vastly outperforms it by reaching $72.8\%$, marking a $5.9\%\uparrow$ advantage. Similarly, when compared against the supervised FrameOracle~\cite{li2025frameoracle} on Qwen3-VL, ReMem achieves $77.6\%$ compared to the baseline's $62.9\%$ on MLVU~\cite{zhou2024mlvu}, yielding a $14.7\%\uparrow$ edge. This confirms that explicitly modeling visual-semantic-temporal correlations mathematically is more efficient and accurate than implicitly learning them through brute-force training.

\noindent\textbf{Robustness Across Diverse Temporal Granularities.} 
Our method's superiority is vividly demonstrated on the Video-MME~\cite{fu2025video} benchmark, containing Short, Medium, and Long subsets (1.3 to 41 minutes). As video length extends, existing methods suffer severe information degradation. While AKS~\cite{tang2025adaptive} enhances LLaVA-Video on Short videos, it stagnates on the Long subset at $54.1\%$. In contrast, our approach maintains semantic coherence, achieving a remarkable $61.8\%$ on Long videos ($7.7\%\uparrow$ edge over AKS~\cite{tang2025adaptive}). This robust trend continues with Qwen2-VL: whereas Q-Frame~\cite{zhang2025q} degrades to $57.1\%$ on extended durations, our framework outperforms it by securing $65.7\%$. These gains prove our dynamic allocation successfully bridges the semantic gap in long-range reasoning.

\subsection{Ablation Studies}

\begin{table*}[t]
\centering
\setlength{\tabcolsep}{5pt} 
\caption{Ablation study across multiple benchmarks with LLaVA-Video model.}
\label{tab:ablation_llava}
\begin{tabular}{cccc|cccc}
\toprule
\multicolumn{4}{c|}{Components} & \multirow{2}{*}{LVBench} & \multirow{2}{*}{MLVU} & \multirow{2}{*}{LongVideoBench} & \multirow{2}{*}{Video-MME} \\
EE & VSA & TSA & FR & & & & \\
\midrule
$\times$ & \checkmark & \checkmark & \checkmark & $52.3_{\color{ForestGreen}{\uparrow 2.2}}$ & $75.8_{\color{ForestGreen}{\uparrow 1.5}}$ & $66.7_{\color{ForestGreen}{\uparrow 0.4}}$ & $68.1_{\color{ForestGreen}{\uparrow 1.1}}$ \\
\checkmark & $\times$ & \checkmark & \checkmark & $41.6_{\color{ForestGreen}{\uparrow 12.9}}$ & $64.0_{\color{ForestGreen}{\uparrow 13.3}}$ & $57.4_{\color{ForestGreen}{\uparrow 9.7}}$ & $60.9_{\color{ForestGreen}{\uparrow 8.3}}$ \\
\checkmark & \checkmark & $\times$ & \checkmark & $47.6_{\color{ForestGreen}{\uparrow 6.9}}$ & $72.9_{\color{ForestGreen}{\uparrow 4.4}}$ & $61.5_{\color{ForestGreen}{\uparrow 5.6}}$ & $64.8_{\color{ForestGreen}{\uparrow 4.4}}$ \\
\checkmark & \checkmark & \checkmark & $\times$   & $51.9_{\color{ForestGreen}{\uparrow 2.6}}$ & $73.6_{\color{ForestGreen}{\uparrow 3.7}}$ & $61.7_{\color{ForestGreen}{\uparrow 5.4}}$ & $62.3_{\color{ForestGreen}{\uparrow 6.9}}$ \\ 
\checkmark & \checkmark & \checkmark & \checkmark & \textbf{54.5} & \textbf{77.3} & \textbf{67.1} & \textbf{69.2} \\
\bottomrule
\end{tabular}
\end{table*}

\begin{figure}[t]
    \centering
    \begin{subfigure}[b]{0.48\textwidth}
        \centering
        \includegraphics[width=\textwidth]{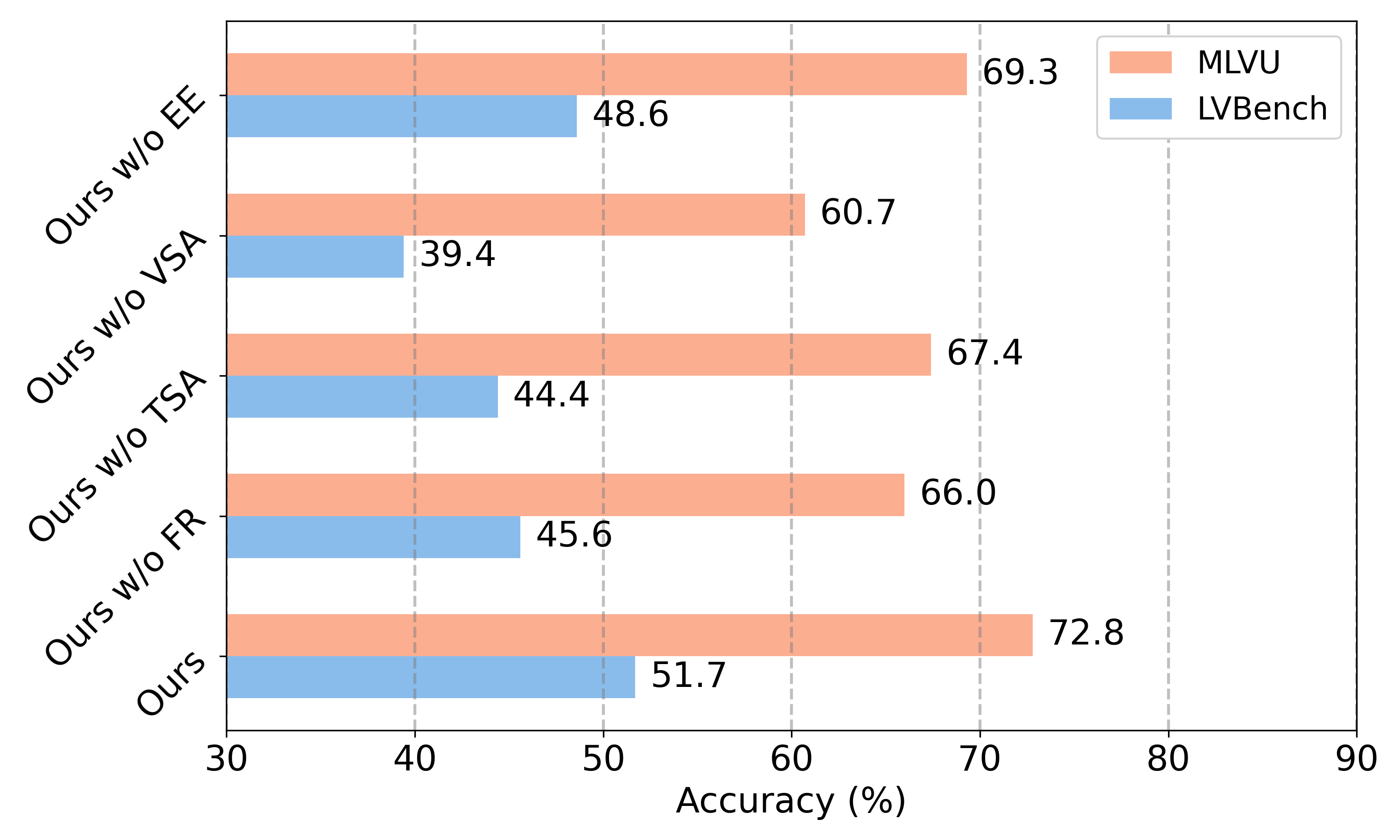} 
        \caption{Ablation study on components across MLVU and LVBench.}
        \label{fig:ablation_all}
    \end{subfigure}
    \hfill 
    \begin{subfigure}[b]{0.48\textwidth}
        \centering
        \includegraphics[width=\textwidth]{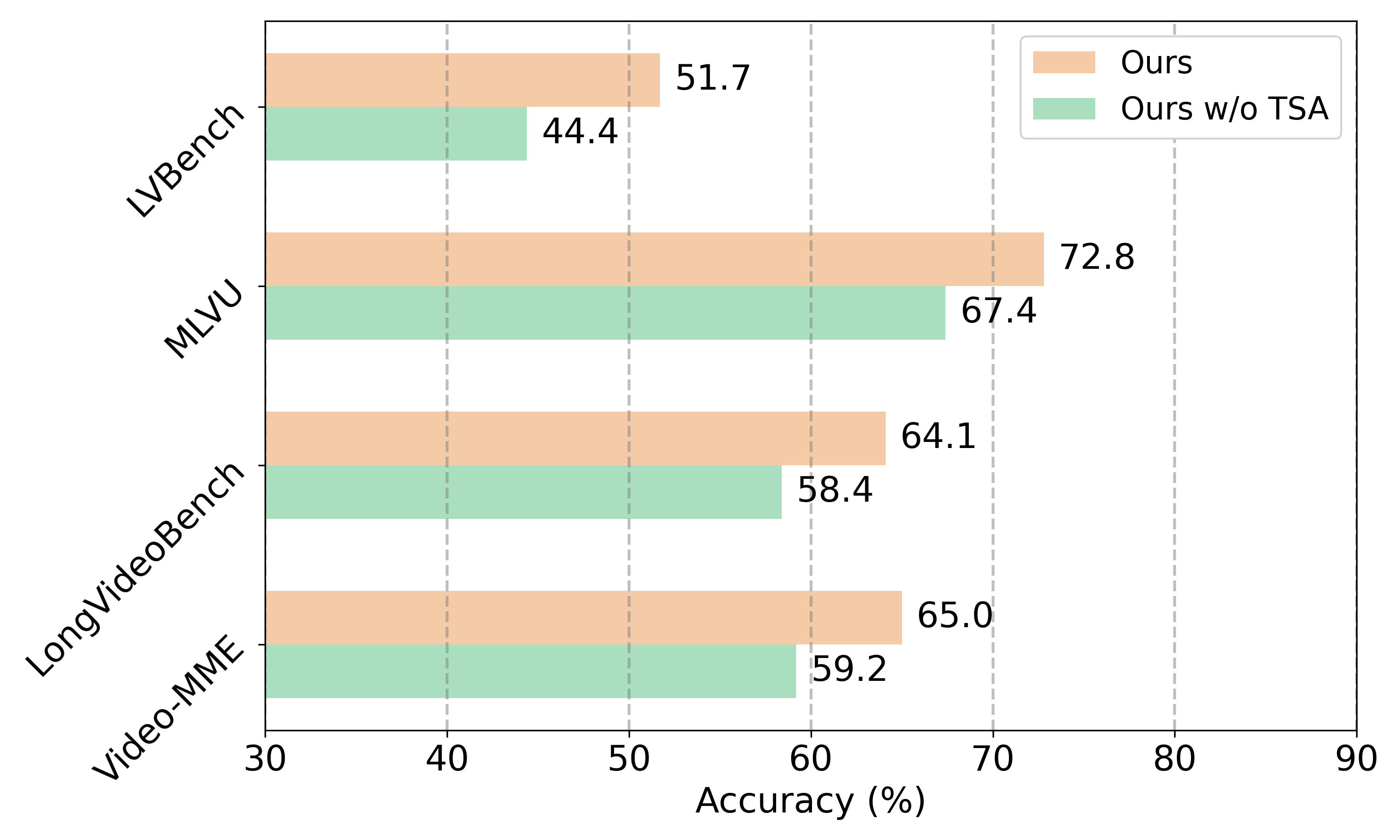} 
        \caption{Ablation study on TSA across multiple benchmarks.}
        \label{fig:ablation_TSA}
    \end{subfigure}
    
    \caption{Ablation results of Qwen2-VL using ReMem. (a) compares components of our proposed framework on Qwen video model. (b) details the impact of removing TSA.}
    \label{fig:ablation_qwen2vl}
    \vspace{-1.2em}
\end{figure}

To evaluate the contribution of each component in our proposed ReMem framework, we conduct extensive ablation studies. We primarily ablate four components: (1) whether \textit{Entity Extraction} (EE) is applied, (2) whether \textit{Static Visual-Semantic Alignment} (VSA) is used, (3) whether \textit{Memory Augmented Temporal-Semantic Alignment} (TSA) is engaged, and (4) whether \textit{Structure-Aware Dynamic Frame Routing} (FR) is enabled. To demonstrate the model-agnostic nature of ReMem, experiments are conducted using both LLaVA-Video (Table \ref{tab:ablation_llava}) and Qwen2-VL (Fig.~\ref{fig:ablation_qwen2vl}) across multiple benchmarks. The results clearly indicate that all components are indispensable and act synergistically.

\noindent\textbf{Effect of Entity Extraction.}
As demonstrated in Table \ref{tab:ablation_llava}, incorporating Entity Extraction yields persistent performance gains across all datasets. Notably, EE integration brings solid improvements on challenging benchmarks, securing a $2.2\%\uparrow$ on LVBench~\cite{wang2025lvbench} and a $1.5\%\uparrow$ on MLVU~\cite{zhou2024mlvu}. Extracting entities from the original query provides fine-grained contextual focus, proving that explicit language priors are highly beneficial for guiding memory graph construction with correct objects and actions.

\noindent\textbf{Static Visual-Semantics Alignment.}
As co-validated by Table \ref{tab:ablation_llava} and Fig.~\ref{fig:ablation_all}, equipping the framework with VSA triggers the most dramatic performance surges across different LLM architectures, proving its role as the core grounding mechanism. For instance, on LVBench~\cite{wang2025lvbench}, integrating VSA recovers a massive 12.9\% accuracy for LLaVA-Video, alongside a mirrored $12.3\%\uparrow$ for Qwen2-VL. This consistent, cross-model boost demonstrates that VSA successfully establishes the visual-textual bridge, enabling the agent to accurately capture specific visual evidence rather than reasoning blindly.

\noindent\textbf{Memory Augmented Temporal-Semantics Alignment.}
The critical role of TSA is evident in Table \ref{tab:ablation_llava}, demonstrating consistent accuracy gains across all four benchmarks upon its integration. Further investigated in Fig.~\ref{fig:ablation_TSA}, integrating TSA into Qwen2-VL drives striking improvements across the board: an absolute gain of 5.8\% on Video-MME~\cite{fu2025video}, 5.4\% on MLVU~\cite{zhou2024mlvu}, and a remarkable $7.3\%\uparrow$ on LVBench~\cite{wang2025lvbench}. This confirms that relying solely on isolated, static visual alignments is inadequate for long-form contexts. By encoding relative temporal distances and cross-frame correlations, TSA bridges visually divergent but causally linked events across extended time spans, preserving the intrinsic temporal logic of the video.

\noindent\textbf{Structure-Aware Dynamic Frame Routing.} Absolute score rankings inherently cause context redundancy by isolating single dominant events according to Table \ref{tab:ablation_llava} and Fig.~\ref{fig:ablation_all}. Replacing this Top-$K$ approach with FR drives a $6.8\%\uparrow$ accuracy gain on MLVU~\cite{zhou2024mlvu} for Qwen2-VL and consistent improvements for LLaVA-Video. This confirms that our temporal granularity-adaptive routing successfully filters redundancy, retaining visual tokens with optimal semantic relevance and temporal diversity.

\subsection{Candidate Pool Capacity Scaling}

\begin{table}[t]
\centering
\setlength{\tabcolsep}{12pt}
\caption{Performance of LLaVA-Video model with ReMem on Video-MME and LongVideoBench under varying
candidate pool sizes. We report accuracy and preprocessing latency (frame selection duration per item).}
\label{tab:frame_num}
\begin{tabular}{@{}lcccc@{}}
\toprule
\multirow{2}{*}{Frames} & \multicolumn{2}{c}{Video-MME} & \multicolumn{2}{c}{LongVideoBench} \\ \cmidrule(l){2-3} \cmidrule(l){4-5} 
 & Acc(\%) & Time(s) & Acc(\%) & Time(s) \\ \midrule
90 & $65.5_{\color{ForestGreen}{\uparrow 3.7}}$ & $17.3_{\color{Red}{\uparrow 1.4}}$ & $62.0_{\color{ForestGreen}{\uparrow 5.1}}$ & $13.6_{\color{Red}{\uparrow 1.6}}$ \\
120 & $67.9_{\color{ForestGreen}{\uparrow 1.3}}$ & $17.5_{\color{Red}{\uparrow 1.2}}$ & $64.9_{\color{ForestGreen}{\uparrow 2.2}}$ & $14.5_{\color{Red}{\uparrow 0.7}}$ \\
\textbf{150} & \textbf{69.2} & \textbf{18.7} & \textbf{67.1} & \textbf{15.2} \\
180 & $68.5_{\color{ForestGreen}{\uparrow 0.7}}$ & $20.2_{\color{ForestGreen}{\downarrow 1.5}}$ & $65.7_{\color{ForestGreen}{\uparrow 1.4}}$ & $16.0_{\color{ForestGreen}{\downarrow 0.8}}$ \\
200 & $68.9_{\color{ForestGreen}{\uparrow 0.3}}$ & $20.9_{\color{ForestGreen}{\downarrow 2.2}}$ & $64.6_{\color{ForestGreen}{\uparrow 2.5}}$ & $16.6_{\color{ForestGreen}{\downarrow 1.4}}$ \\ \bottomrule
\end{tabular}
\end{table}

To determine the optimal size $N$ of the candidate pool prior to \textit{Structure-Aware Dynamic Frame Routing}, we evaluate the trade-off between reasoning accuracy and preprocessing latency. Since the final budget of frames fed into MLLMs remains fixed as $B$, $N$ strictly controls the sampling capacity. 

As illustrated in Table \ref{tab:frame_num}, expanding the candidate pool initially drives steady performance gains. The accuracy climbs as $N$ increases from 90 to 150, peaking at 69.2\% and 67.1\% respectively, confirming that a broad candidate pool prevents the premature pruning of vital visual evidence. However, expanding the pool further triggers a noticeable accuracy drop alongside a linear surge in preprocessing time (e.g., $N=200$ drops accuracy to 68.9\% on VideoMME~\cite{fu2025video}). This exposes a characteristic of clustering-based condensation: an excessively large pool introduces a `long tail' of noisy, low-relevance frames. These redundant frames severely distort the clustering feature space, diluting the semantic density of the final condensed budget. Consequently, $N=150$ emerges as the optimal size, balancing sufficient diversity with required purity.

\begin{figure}[t]
    \centering
    \begin{minipage}[b]{0.48\textwidth}
        \centering
        \includegraphics[width=\textwidth]{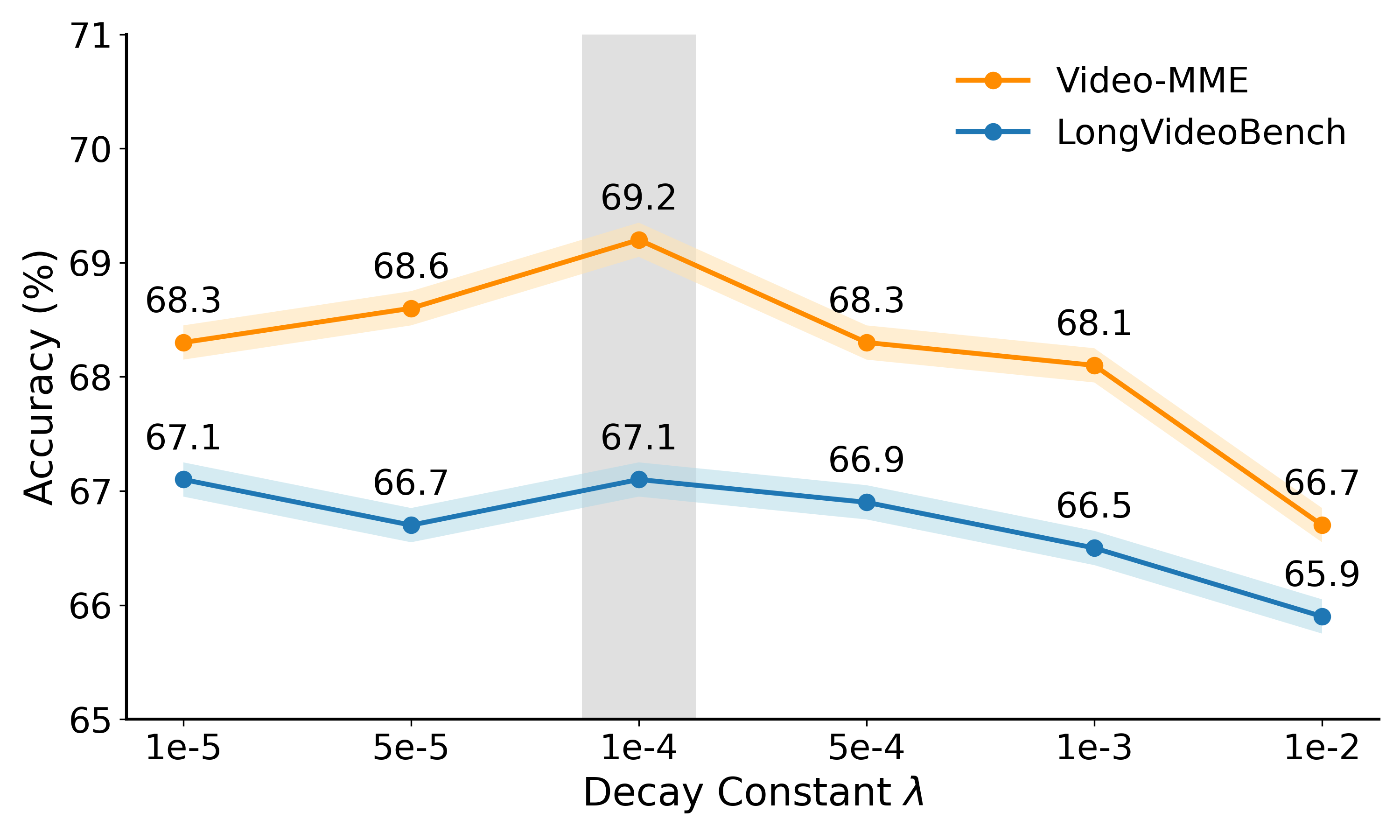}
        \caption{Model performance on two benchmarks with varying decay constant.}
        \label{fig:lamda}
    \end{minipage}
    \hfill 
    \begin{minipage}[b]{0.48\textwidth}
        \centering
        \includegraphics[width=\textwidth]{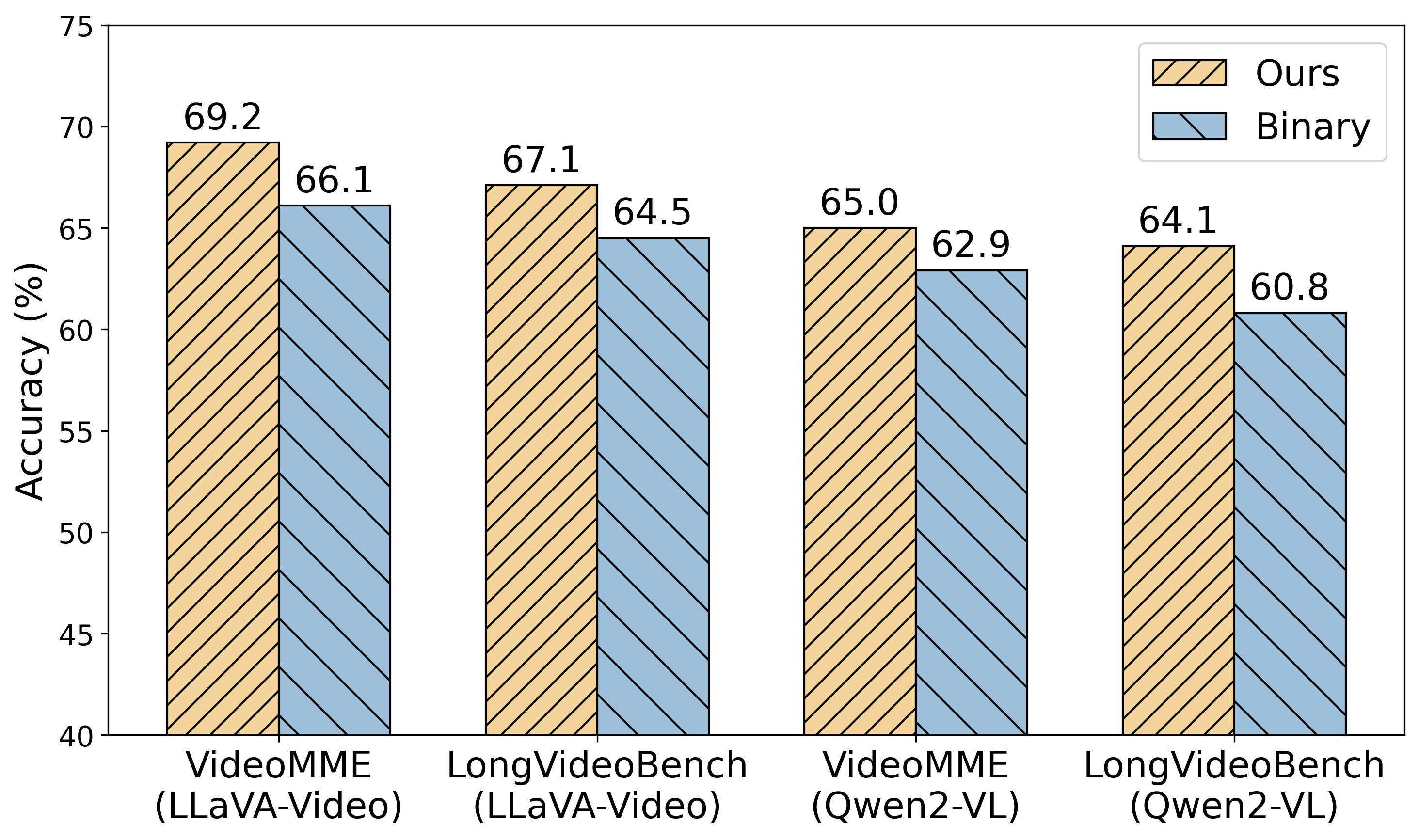}
        \caption{Performance comparison of different query temporal granularity analysis.}
        \label{fig:time_granularity}
    \end{minipage}
    \vspace{-1.2em}
\end{figure}

\subsection{Sensitivity Analysis of Decay Constant}

To determine the optimal decay constant $\lambda$, we analyze its impact on reasoning accuracy. As illustrated in Fig.~\ref{fig:lamda}, accuracy on both VideoMME~\cite{fu2025video} and LongVideoBench~\cite{wu2024longvideobench} exhibits a clear inverted-V trend as $\lambda$ increases from $10^{-5}$ to $10^{-2}$. Performance peaks at $\lambda = 10^{-4}$ (achieving 69.2\% and 67.1\% respectively) and drops significantly as $\lambda$ grows larger. 
This trajectory directly reflects the trade-off governed by $\lambda$. When $\lambda$ is excessively small, the decay mechanism is sluggish. The model over-relies on static visual evidence for long videos, hindering the timely injection of temporal-semantic alignment. Conversely, as $\lambda$ increases to a larger value, premature decay forces $\alpha$ to 0.5 too early, neutralizing the temporal granularity $g$ and preventing the model from adequately prioritizing instantaneous spatial cues. Therefore, $\lambda = 10^{-4}$ provides the optimal balance, ensuring a well-calibrated transition from localized visual focus to global temporal reasoning.

\subsection{Effect of Dynamic Time Granularity}

To validate the necessity of the continuous Time Granularity, we compare our dynamic weighting mechanism $\mathcal{S} = \alpha \mathcal{S}_{\text{v}} + (1 - \alpha) \mathcal{S}_{\text{t}}$ against rigid binary classification (short-term $\alpha = 0.9$, long-term $\alpha = 0.5$). 
As illustrated in Fig.~\ref{fig:time_granularity}, upgrading to our dynamic approach yields consistent and notable performance gains across both LLM architectures. For instance, on LongVideoBench~\cite{wu2024longvideobench}, our dynamic method boosts Qwen2-VL accuracy by 3.3\% and LLaVA-Video by 2.6\%. This performance gap exposes a fundamental limitation of the binary approach: long videos encapsulate complex events spanning a continuous spectrum of temporal scales. A simple dichotomy lacks the necessary resolution to capture these subtleties. Our dynamic weighting gracefully adapts to diverse granularities, establishing a precise alignment between the query's temporal demands and the video's topological memory.

\subsection{Efficiency Analysis and Trade-off Discussion}

\begin{figure}[t]
    \centering
    \includegraphics[width=0.48\textwidth]{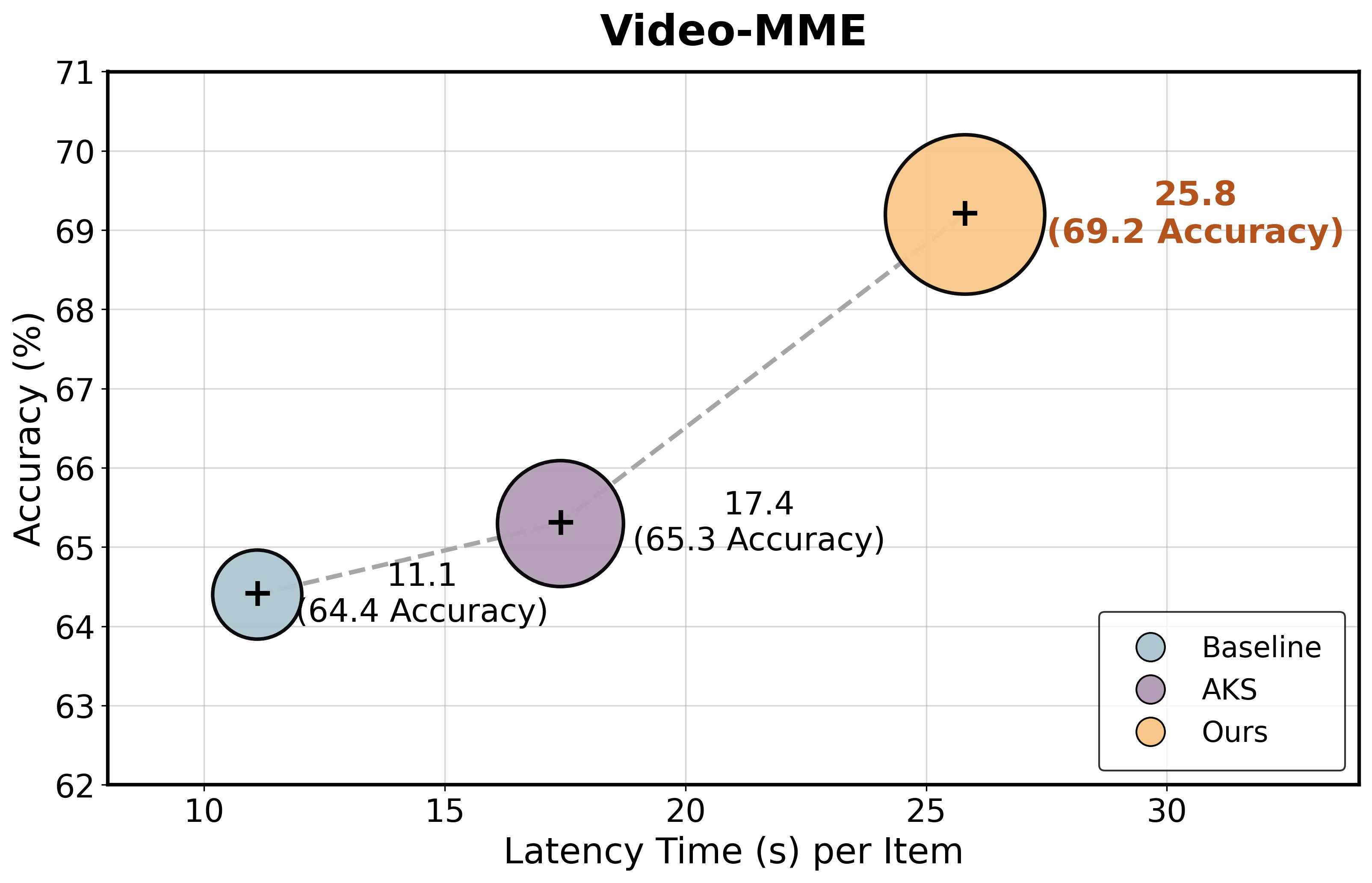}
    \hfill 
    \includegraphics[width=0.48\textwidth]{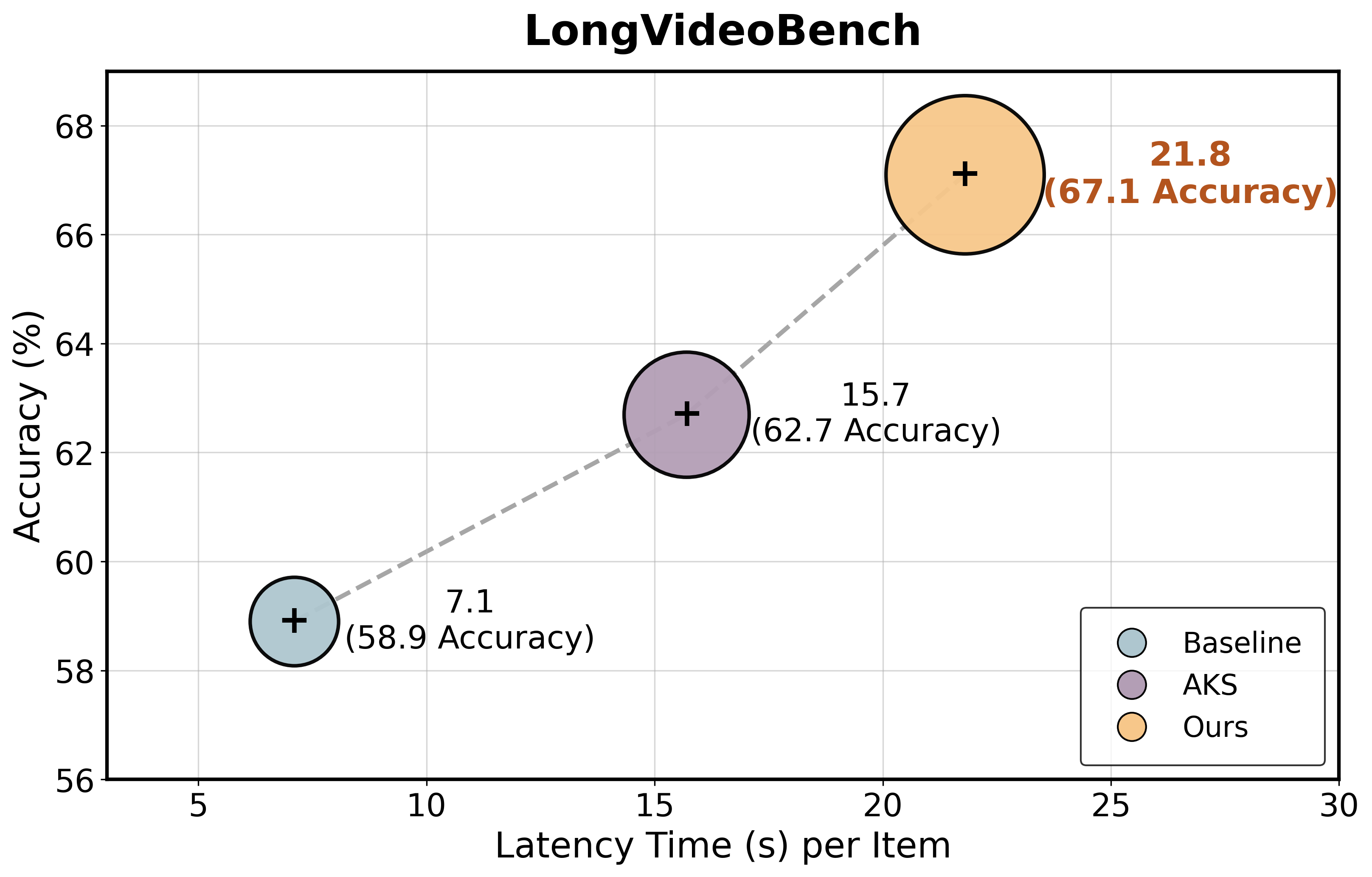}
    
    \caption{Visualization of LLaVA-Video reasoning accuracy and latency time with different frame selection methods on Video-MME (left) and LongVideoBench (right).}
    \label{fig:efficiency}
    \vspace{-1.2em}
\end{figure}

As illustrated in Fig.~\ref{fig:efficiency}, our method achieves state-of-the-art accuracy with slightly higher end-to-end inference time per sample compared to Uniform Sampling (used by baseline) and AKS~\cite{tang2025adaptive}. However, this increase in total latency is a deliberate and highly rewarding accuracy-efficiency trade-off.

\noindent\textbf{Favorable Marginal Efficiency.}
The increased latency primarily stems from the keyframe selection phase, where our module delves deeper into the video to capture richer spatial-temporal information. Importantly, this extra time investment yields exceptionally high returns. Observing the Pareto frontier in Fig.~\ref{fig:efficiency}, our method demonstrates superior efficiency: on Video-MME~\cite{fu2025video}, while AKS~\cite{tang2025adaptive} improves baseline accuracy by $0.9\%$ with \(+6.3s\), our method yields a substantial $3.9\%\uparrow$ over AKS~\cite{tang2025adaptive} with \(+8.4s\), highlighting a much more effective utilization of computational resources.

\noindent\textbf{Amortized Cost via Plug-and-Play Design.} 
Our keyframe selection module functions as a flexible, model-agnostic plug-and-play interface. When processing the same benchmark, keyframe indices can be directly reused for inference across different backbones. This significantly amortizes the selection time, making our pipeline highly efficient for multi-model testing scenarios.

\subsection{Qualitative Analysis}

\begin{figure}[tb]
  \centering
  \includegraphics[height=8cm]{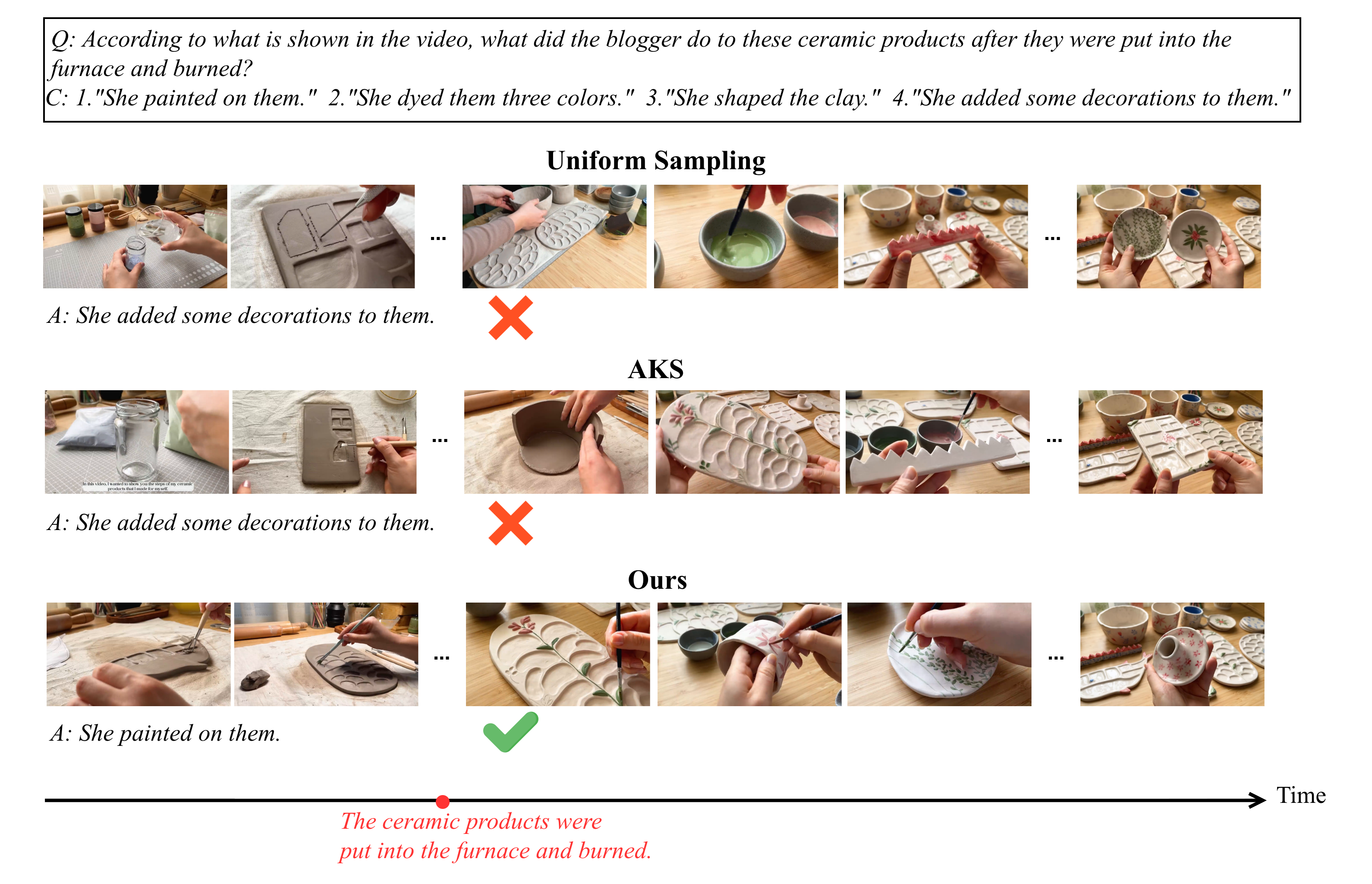}
  \caption{Case analysis from Video-MME. Unlike Uniform Sampling and AKS which miss crucial temporal context, our method accurately routes the relevant frames following the key event (red dot) to answer correctly.}
  \label{fig:qualitative}
  \vspace{-1.2em}
\end{figure}

To intuitively demonstrate the superiority of our frame selection strategy, we present a qualitative comparison in Fig.~\ref{fig:qualitative}, visualizing frames sampled by Uniform Sampling, AKS~\cite{tang2025adaptive}, and our ReMem for a complex QA case. The query is highly challenging as it demands explicit temporal logic reasoning: the model must temporally localize a milestone and exclusively attend to subsequent actions, filtering out prior events.

As shown, both Uniform Sampling and AKS~\cite{tang2025adaptive} fail. Uniform Sampling rigidly distributes frames across the entire video, allocating too much attention to early stages and diluting critical post-furnace evidence. Similarly, AKS~\cite{tang2025adaptive} retrieves visually dynamic but temporally irrelevant frames from the preliminary phase. Conversely, ReMem bypasses distracting early phases and densely samples the critical frames after the furnace milestone, which explicitly depict the blogger painting the fired ceramics. This compellingly validates our method's ability to isolate crucial, query-aligned visual cues for complex temporal reasoning.

\begin{figure}[tb]
  \centering
  \includegraphics[width=\linewidth]{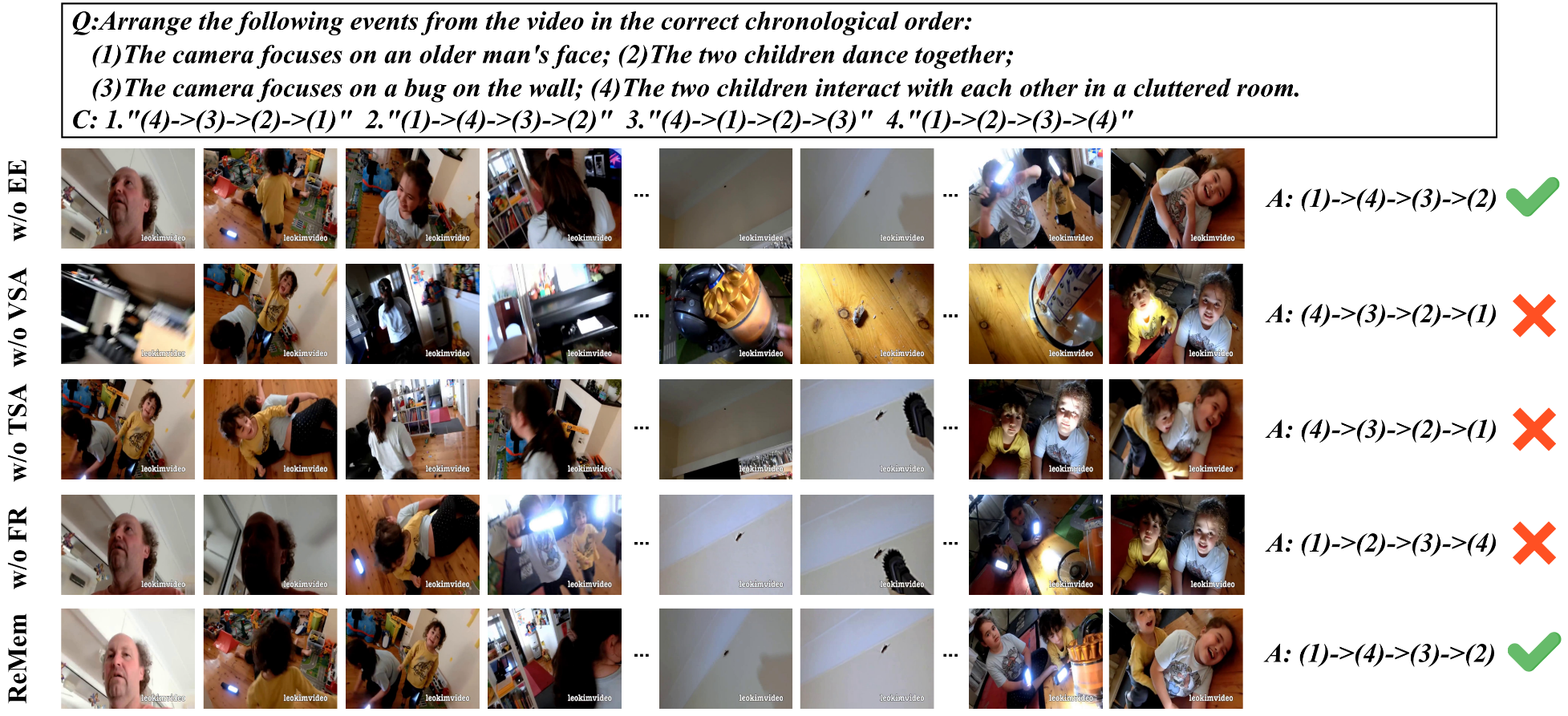}
  \caption{Case analysis from MLVU. Variants without key components of our proposed ReMem fail.}
  \label{fig:qualitative_ablation}
  \vspace{-1.2em}
\end{figure}

To complement the quantitative findings in ablation studies, Fig.~\ref{fig:qualitative_ablation} qualitatively illustrates the necessity of each ReMem module. As depicted, removing EE does not impact the outcome in this specific instance, as the candidate pool already provides sufficient information. However, dropping VSA or TSA yields an incomplete temporal span, missing the initial appearance of the older man's face. Meanwhile, the variant without FR manages to find the correct visual cues but disrupts temporal coherence, leading to a scrambled event sequence. Ultimately, the complete ReMem pipeline precisely extracts frames across the timeline, depicting all four anchor events in correct chronological order. This visualization underscores how our modules work synergistically to guarantee both visual-semantic relevance and temporal structural integrity.

\vspace{-1em}
\section{Conclusion}

In this work, we presented \textbf{ReMem}, a training-free, temporal granularity-adaptive keyframe selection framework to enhance zero-shot LongVideoQA. By introducing a memory-augmented dual-level adaptation, ReMem effectively integrates \emph{query-level adaptation} through Memory-Driven Question Parsing and \emph{video-level adaptation} via Synergistic Dual-Semantic Frame Alignment and Structure-Aware Dynamic Frame Routing.
This synergistic combination allows the framework to identify the most discriminative frames while explicitly preserving causal structures and temporal information tailored to the varying query temporal granularities.
Extensive evaluations across four LongVideoQA benchmarks demonstrate that ReMem significantly improves zero-shot reasoning capabilities across multiple MLLMs, achieving state-of-the-art performance without any parameter tuning. Our results highlight the potential of unified memory mechanisms as a powerful tool for robust long video understanding, opening avenues for future research in efficient and scalable training-free spatiotemporal reasoning.

\section*{Acknowledgements}
This work was supported by Ministry of Education Tier 2 grant, Singapore (T2EP20224-0028), and Ministry of Education Tier 1 grant, Singapore (23-0651-P0001). 

%
%
\bibliographystyle{splncs04}
\bibliography{main}
\end{document}